\documentclass[letterpaper,conference]{IEEEtran}
\IEEEoverridecommandlockouts
\usepackage{graphics} 
\usepackage[linesnumbered,ruled]{algorithm2e}
\graphicspath{{img/}}
\usepackage{epsfig} 
\usepackage{amsmath} 
\usepackage{textcomp}
\usepackage{amssymb}  
\usepackage{amsthm}
\usepackage{bigints}
\usepackage{pifont}
\def\-{\raisebox{.75pt}{-}}
\usepackage{mathtools}
\usepackage{cite}
\usepackage[makeroom]{cancel}
\usepackage{xcolor} 

\usepackage[center]{subfigure}
\usepackage{multirow}
\usepackage{booktabs}
\usepackage{subfigure}
\usepackage{epstopdf}
\usepackage{textcomp}
\usepackage{hyperref}
\usepackage{bm}
\usepackage{caption}
\usepackage{algpseudocode,algorithm} 
\usepackage{color}
\usepackage{mathptmx}
\usepackage[11pt]{moresize}
\usepackage{amsmath}
\usepackage{hhline}
\usepackage{tikz}
\usepackage{lipsum}

\usepackage{textcomp}
\def\BibTeX{{\rm B\kern-.05em{\sc i\kern-.025em b}\kern-.08em
    T\kern-.1667em\lower.7ex\hbox{E}\kern-.125emX}}

\newcommand\copyrighttext{%
  \centering \footnotesize \copyright 2018 IEEE, to appear Proceedings of the 21st IEEE International Conference on Intelligent Transportation Systems}
\newcommand\copyrightnotice{%
\begin{tikzpicture}[remember picture,overlay]
\node[anchor=south,yshift=10pt] at (current page.south) {\fbox{\parbox{\dimexpr\textwidth-\fboxsep-\fboxrule\relax}{\copyrighttext}}};
\end{tikzpicture}%
}

\begin{document}
\title{\LARGE Towards Safe Autonomous Driving: Capture Uncertainty in the Deep Neural Network For Lidar 3D Vehicle Detection
\thanks{$^1$ Di Feng, Lars Rosenbaum are with Robert Bosch GmbH, Corporate Research, Driver Assistance Systems and Automated Driving, 71272 Renningen, Germany.}
\thanks{$^2$ Klaus Dietmayer is with Institute of Measurement, Control and Microtechnology, Ulm University, 89081 Ulm, Germany.}
\thanks{The video to this paper can be found at \url{https://youtu.be/bQJmssB80oM}.}
}

\author{Di Feng$^1$, Lars Rosenbaum$^1$, Klaus Dietmayer$^2$}

\maketitle

\begin{abstract}
To assure that an autonomous car is driving safely on public roads, its object detection module should not only work correctly, but show its prediction confidence as well. Previous object detectors driven by deep learning do not explicitly model uncertainties in the neural network. We tackle with this problem by presenting practical methods to capture uncertainties in a 3D vehicle detector for Lidar point clouds. The proposed probabilistic detector represents reliable \textit{epistemic} uncertainty and \textit{aleatoric} uncertainty in classification and localization tasks. Experimental results show that the epistemic uncertainty is related to the detection accuracy, whereas the aleatoric uncertainty is influenced by vehicle distance and occlusion. The results also show that we can improve the detection performance by $1\%-5\%$ by modeling the aleatoric uncertainty.  
\end{abstract}

\copyrightnotice

\section{\textbf{Introduction}}\label{sec:intro}
Knowing what an object detection model is unsure about is of paramount importance for safe autonomous driving. For example, if an autonomous car recognizes a front object as a pedestrian but is uncertain about its location, the system may warn the driver to take over the car at an early stage or slow down to avoid fatal accidents. 

Deep learning has been introduced to object detection in the autonomous driving settings that use cameras \cite{chen2016monocular,chen20153d,mousavian20173d}, Lidar \cite{li20163d,li2016vehicle,zhou2017voxelnet,engelcke2017vote3deep,caltagirone2017fast,asvadi2017depthcn,yang2018pixor}, or both \cite{chen2016multi,qi2017frustum,xu2017pointfusion,ku2017joint,wang2017fusing,du2018general,pfeuffer2018optimal}, and has set the benchmark on many popular datasets (e.g. KITTI \cite{Geiger2013IJRR}, Cityscapes \cite{Cordts2016Cityscapes}). However, to the best of our knowledge, none of these methods allow for the estimation of uncertainty in bounding box regression. Moreover, these methods often use a softmax layer to normalize the score vector on the purpose of classifying objects, which does not necessarily represent the classification uncertainty in the model. As a result, these object detections can only tell the human drivers \textit{what} they have seen, but not \textit{how certain} they are about it.
\begin{figure}[htbp]
	\centering
\includegraphics[width=0.7\linewidth]{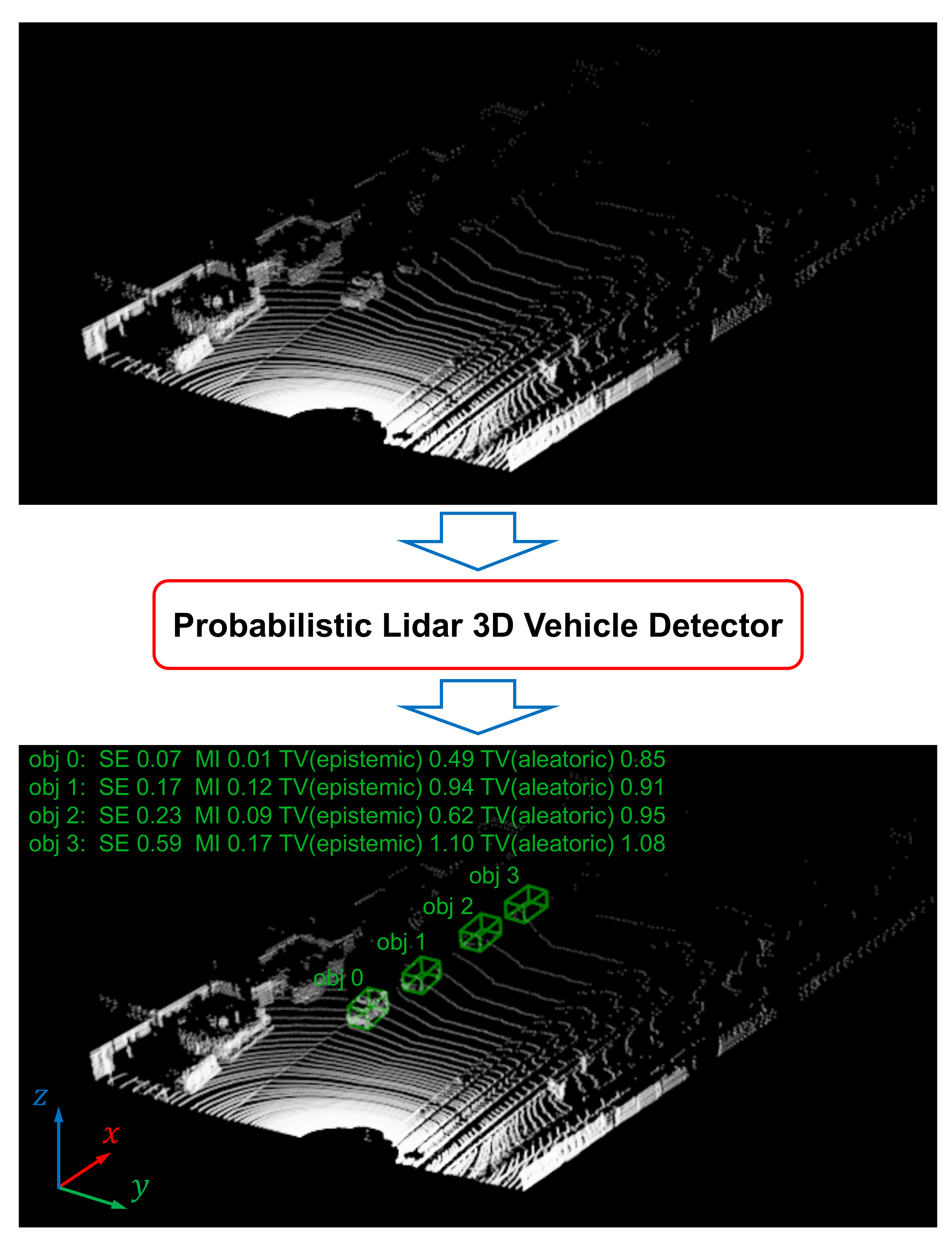}
\caption{Our proposed probabilistic Lidar 3D vehicle detection network takes the Lidar point clouds as input. It not only predicts object classes and 3D bounding boxes, but also predicts the model uncertainty and the sensor observation uncertainty. Shannon Entropy (SE) and Mutual Information (MI) quantify the classification uncertainty, and Total Variance (TV) the localization uncertainty. These scores will be described in Sec.~\ref{sec:bnn}.}\label{fig:introduction}
\vspace{-1.2em}
\end{figure}

There are two types of uncertainty that we can quantify in the object detection network. Epistemic uncertainty, or model uncertainty, indicates how uncertain an object detector is to explain the observed dataset. Aleatoric uncertainty, conversely, captures the observation noises that are inherent in sensors. For instance, detecting an abnormal object that is different from the training dataset may result in high epistemic uncertainty, while detecting a distant object may result in high aleatoric uncertainty. Capturing both uncertainties in the object detection network is indispensable for safe autonomous driving, as the epistemic uncertainty displays the limitation of detection models, while the aleatoric uncertainty can provide the sensor observation noises for object tracking.

In this work, we develop practical methods to capture epistemic and aleatoric uncertainties in a 3D vehicle detector for Lidar point clouds. Our contributions are three-fold:
\begin{itemize}
\item We extract model uncertainty and observation uncertainty for the vehicle recognition and 3D bounding box regression tasks. 
\item We show an improvement of vehicle detection performance by modeling the aleatoric uncertainty.
\item We study the difference between the epistemic and aleatoric uncertainty. The former is associated with the vehicle detection accuracy, while the latter is influenced by the vehicle distance and occlusion.
\end{itemize}

The remainder of the paper is structured as follows: Sec.~\ref{sec:related_works} summarizes related works. Sec.~\ref{sec:method} illustrates the architecture of our probabilistic Lidar 3D vehicle detection neural network. Sec.~\ref{sec:bnn} presents the proposed methods to capture epistemic and aleatoric uncertainties. Sec.~\ref{sec:results} illustrates the experimental results, followed by a conclusion and a discussion of future research in Sec.~\ref{sec:conclusion}.

\section{\textbf{Related Works}}\label{sec:related_works}
In this section, we first summarize methods for 3D object detection in autonomous driving for Lidar point clouds. We then summarize related works to Bayesian neural network which we use to extract uncertainty in our Lidar vehicle detection network.
\subsection{3D Object Detection in Autonomous Driving}
\subsubsection{3D Object Detection via Lidar Point Clouds}
Many works represent 3D point clouds using voxel grids. Li \cite{li20163d} discretizes point clouds into square grids, and then employs a 3D convolution neural network that outputs an objectness map and a bounding box map for each grid. Zhou \textit{et al.} \cite{zhou2017voxelnet} introduce a voxel feature encoding (VFE) layer that can learn the unified features directly from the Lidar point clouds. They feed these features to a Region Proposal Network (RPN) that predicts the detections. Engelcke \textit{et al.} \cite{engelcke2017vote3deep} use a voting scheme to search possible object locations, and introduce special convolutional layers to tackle with the sparsity of the point clouds. In addition, several works encode 3D point clouds as 2D feature maps. Li \textit{et al.} \cite{li2016vehicle} project range scans into a 2D front-view depth map and use a 2D fully convolutional network (FCN) to detect vehicles. Caltagirone \textit{et al.} \cite{caltagirone2017fast} and Yang \textit{et al.} \cite{yang2018pixor} project the Lidar point clouds into birds' eye view image and propose a road detector and a car detector, respectively.

\subsubsection{3D Object Detection by Combining Lidar and Camera}
Lidar point clouds usually give us accurate spatial information. Camera images provide us with object appearances, which is beneficial for object classification. Therefore, it is natural to combine these two sensors \cite{chen2016multi,qi2017frustum,xu2017pointfusion,ku2017joint,wang2017fusing,du2018general,pfeuffer2018optimal}. For example, Chen \textit{et al.} \cite{chen2016multi} propose MV3D, a network that generates region proposals from the bird's eye view Lidar features and then combine the proposals with the regional features from the front view Lidar feature maps and RGB camera images for accurate 3D vehicle detection. Qi \textit{et al.} \cite{qi2017frustum} use RGB camera images to draw 2D object bounding boxes, which build frustums in the Lidar point cloud. Then, they use these frustums for 3D bounding box regressions.

\subsection{Bayesian Neural Networks}
Bayesian neural networks (BNNs) provide us the probabilistic interpretation of neural networks \cite{mackay1992practical}. Instead of placing deterministic weights in the neural network, BNNs assume a prior distribution over them. By inferring the posterior distribution over the model weights, we can extract how uncertain a neural network is with its predictions. Uncertainty estimation methods include variational inference \cite{hinton1993keeping}, sampling technique \cite{graves2011practical}, or ensemble \cite{osband2016deep}. Recently, Y.~Gal \cite{Gal2016Uncertainty} propose a method that captures the uncertainty in BNNs at test time by sampling the network multiple times with dropout. This dropout sampling method has been applied to active learning for cancer diagnosis \cite{gal2017deep}, semantic segmentation \cite{kendall2017uncertainties,kendall2015bayesian} and image object detection for open-set problem \cite{miller2017dropout}.

\section{\textbf{Network Architecture}}\label{sec:method}
\begin{figure*}[htpb]
	\centering
	\begin{minipage}{0.8\textwidth}
		\centering
		\includegraphics[width=1\linewidth]{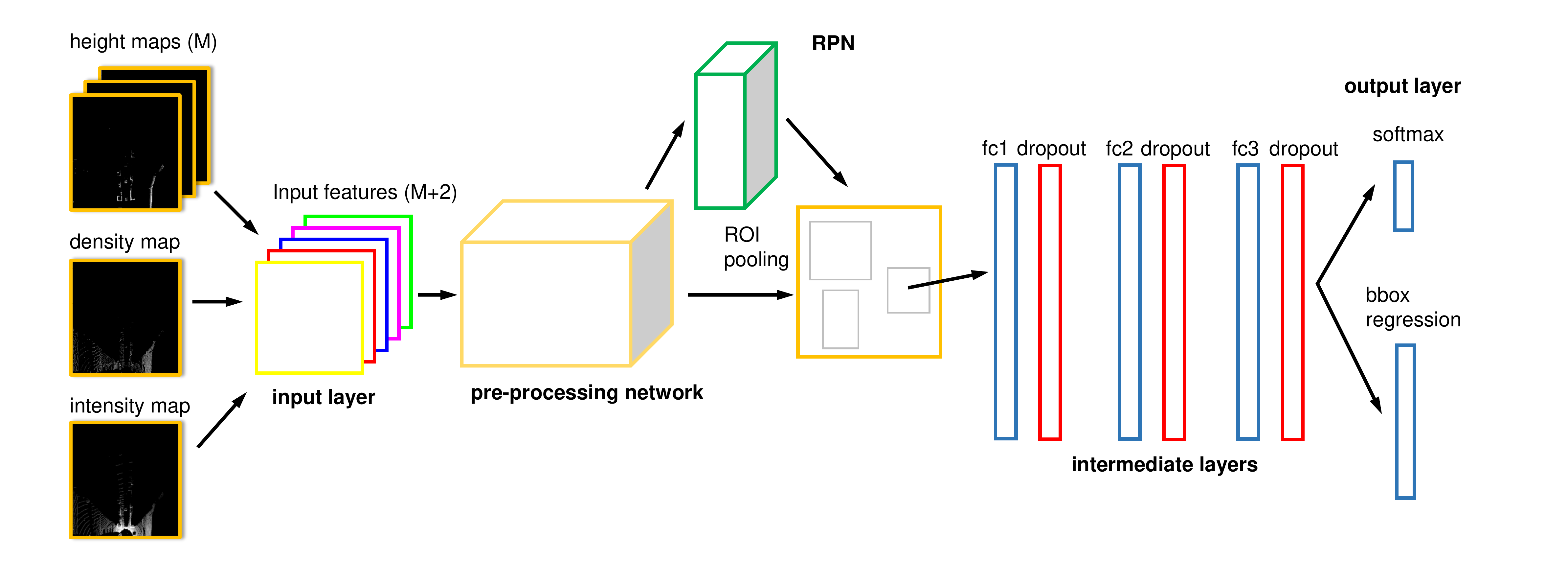}
	\end{minipage}
	\caption{Network architecture for our proposed probabilistic Lidar 3D vehicle detector.}\label{fig:network}
\end{figure*}

Our proposed probabilisitic Lidar 3D object detection network is shown in Fig.~\ref{fig:network}. The network takes the features from Lidar bird's eye view (BEV) as input and feed it into a pre-processing network. Then, a region proposal network is employed to generate candidates of regions of interest (ROIs), which are processed through the intermediate layers. These layers are built to capture the uncertainties in our object detection network. After the intermediate layers, the features are fed into two fully connected layers for $3$D bounding box regression and softmax objectness score.

\subsection{Input and Output Encoding}
To extract the Lidar BEV features, the 3D point clouds are first projected onto a 2D grid with a resolution of $0.1$m and then encoded by height, intensity and density maps \cite{chen2016multi}. The height maps are generated by dividing the point clouds into $M$ slices. As a result we obtain $M+2$ channel features. Fig.~\ref{fig:network} shows exemplary input features.

The network outputs the softmax objectness score and the oriented 3D bounding boxes. The bounding boxes $\mathbf{v}$ are encoded by their $8$ corner offsets, $\mathbf{v} = [x_0,...,x_7; y_0,...,y_7; z_0,...,z_7]$, and are normalized by the diagonal length of the proposal anchors. Concretely, let $\mathbf{\hat{v}}$ be the vector of a bounding box in the Lidar coordinate frame and $\hat{\mathbf{v}}_0$ its corresponding region proposal with diagonal length $h$, we encode the $24$-dimensional regression outputs by $\mathbf{v} = \frac{\mathbf{\hat{v}}-\hat{\mathbf{v}}_0}{h}$.  

\subsection{Feature Pre-processing Network}
\begin{figure}[htbp]
	\centering
		\centering
		\includegraphics[width=0.96\linewidth]{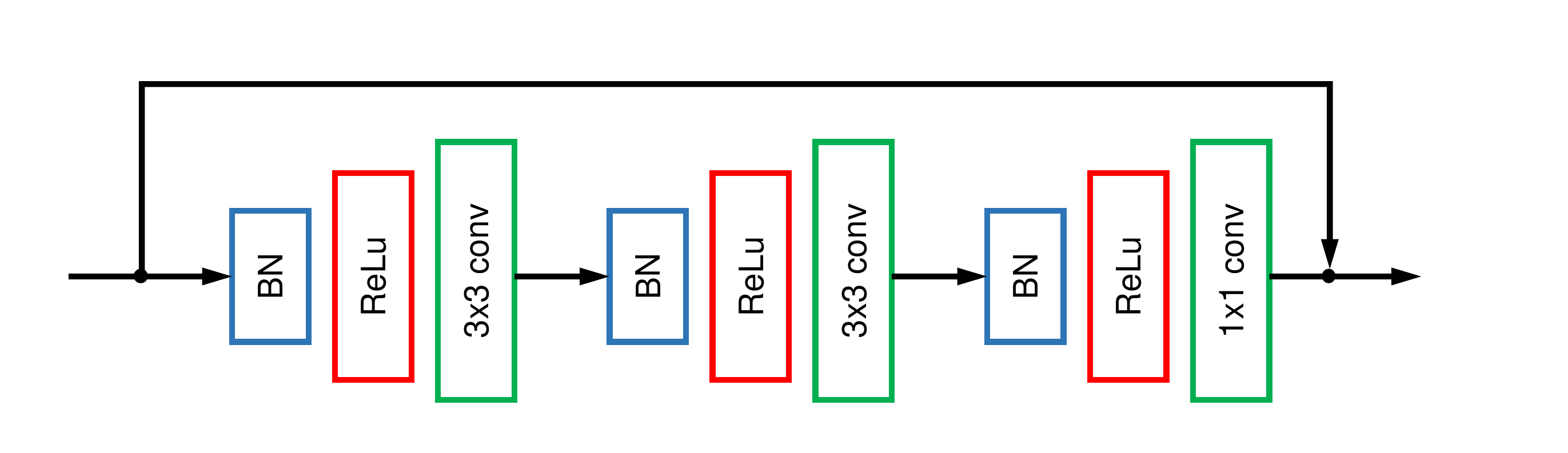}
	\caption{Resnet block architecture.}\label{fig:resnet}
\end{figure}

To process the Lidar feature images, we use the ResNet-8 \cite{he2016identity} architecture that contains $4$ residual blocks (one block architecture is shown in Fig.~\ref{fig:resnet}), with the number of kernels increasing from $64$, $128$, $256$ to $512$. The last convolution layer is up-sampled with a factor of two for generating region proposals and detecting objects. In this way, the network can detect small vehicles that occupy only a few grid cells ($5$-$40$ grid cells \cite{chen2016multi}).

\subsection{Region Proposal Networks}
We follow the Faster-RCNN pipeline\cite{ren2015faster} to generate BEV 3D region proposals. First, for each feature map pixel we generate nine 2D anchors using $3$ scales with box areas of $16^2$, $32^2$ and $48^2$ and $3$ aspect ratios of $1:1$, $1:2$, and $2:1$. Then, the 2D anchors are projected into 3D space by adding a fixed height value, which is selected large enough to enclose all daily vehicles.

\subsection{Intermediate Layers}
To further process the Lidar features and to extract uncertainties, we design $3$ fully connected hidden layers, each of which has $512$ hidden units and is followed by a dropout layer.  

\section{\textbf{Capturing Uncertainty in Lidar 3D Vehicle Detection}} \label{sec:bnn}
As introduced in Sec.~\ref{sec:intro}, there are two different types of uncertainty that can be captured by our proposed probabilistic Lidar 3D vehicle detector: the epistemic uncertainty that describes the uncertainty in the model, and the aleatoric uncertainty that describes the observation noises.

We employ the intermediate layers (Fig.~\ref{fig:network}) as a Bayesian neural network to extract uncertainties. Let us denote $\mathbf{x}^{\ast}$ as a ROI candidate generated by RPN during the test time, and $\mathbf{y}^{\ast}$ the prediction of object labels or bounding boxes. We want to estimate the posterior distribution of the prediction $p(\mathbf{y}^{\ast}|\mathbf{x}^{\ast},\mathbf{X},\mathbf{Y})$, where $\mathbf{X}$ denotes the training dataset and $\mathbf{Y}$ its ground truth. To do this, we marginalize the prediction over the weights $\mathbf{W}$ in the fully connected layers through $p(\mathbf{y}^{\ast}|\mathbf{x}^{\ast},\mathbf{X},\mathbf{Y}) = \bigintsss p(\mathbf{y}^{\ast}|g^{\mathbf{W}}(\mathbf{x}^{\ast}))p(\mathbf{W}|\mathbf{X},\mathbf{Y})d\mathbf{W}$, where $g^{\mathbf{W}}(\mathbf{x}^{\ast})$ denotes the network output. By estimating the weight posterior $p(\mathbf{W}|\mathbf{X},\mathbf{Y})$ and performing Bayesian inference, we can extract epistemic uncertainty; By estimating the observation likelihood $p(\mathbf{y}^{\ast}|g^{\mathbf{W}}(\mathbf{x}^{\ast}))$, we can obtain the aleatoric uncertainty.

\subsection{Capturing Epistemic Uncertainty}
In order to extract the epistemic uncertainty (or model uncertainty) in our vechile detection network, we need to calculate the weight posterior $p(\mathbf{W}|\mathbf{X},\mathbf{Y})$. However, performing the inference analytically is in general intractable, thus approximation techniques are required. Recently, Gal \cite{Gal2016Uncertainty} shows that dropout can be used as approximate inference in a Bayesian neural network: during the training process, performing dropout with stochastic gradient descent is equivalent to optimizing the approximated posterior distribution with Bernoulli distributions over the weights $\mathbf{W}$. Gal \cite{Gal2016Uncertainty} also shows that by performing the network forward passes several times with dropout during the test time, we can get the samples from the model's posterior distribution, and thus obtain the epistemic uncertainty.

We employ the dropout method to capture the model uncertainty in the object detection network. Our network is trained using dropout with a dropout rate $p$. For each region proposal during test time $\mathbf{x}^{\ast}$, it performs $N$ forward passes with dropout. The output of the network contains the softmax score $\{s_{\mathbf{x}^{\ast}}^{i}\}_{i=1}^{N}$ and the bounding box regression $\{\mathbf{v}_{\mathbf{x}^{\ast}}^{i}\}_{i=1}^{N}$, with $s_{\mathbf{x}^{\ast}}^{i}$ and $\mathbf{v}_{\mathbf{x}^{\ast}}^{i}$ being the softmax score and the regression output for the $i^{th}$ forward pass, respectively. In the following, we illustrate how to use these outputs to measure the epistemic uncertainty.

\subsubsection{Extracting Vehicle Probability and Epistemic Classification Uncertainty}
The vehicle probability $p(veh|\mathbf{x}^{\ast})$ - the probability that a sample $\mathbf{x}^{\ast}$ is recognized as a vehicle - is approximated by the mean of softmax score of $N$ forward passes according to
\begin{align}\label{equ:prob}
\begin{split}
p(veh|\mathbf{x}^{\ast}) & = \mathbb{E}_{p(\mathbf{W}|\mathbf{X},\mathbf{Y})} p(veh|\mathbf{x}^{\ast},\mathbf{W}) \\ & \approx \frac{1}{N}\sum_{i=1}^{N} s_{\mathbf{x}^{\ast}}^{i}.
\end{split}
\end{align}
Note that if we set $N=1$ and $p=0$, $p(veh|\mathbf{x}^{\ast})$ is approximated by a softmax score - the point-wise prediction done by most object detection networks. However, as Gal \cite{Gal2016Uncertainty} mentioned, passing the point estimate can underestimate the uncertainty of samples that are different from the training dataset. In order to extract more accurate model uncertainty, the network needs to perform multiple forward passes with dropout (i.e. $N>1, p>0$).

We further use the Shannon entropy (SE) and mutual information (MI) to measure the classification uncertainty \cite{Gal2016Uncertainty}. For a region proposal $\mathbf{x}^{\ast}$, its SE score is calculated via
\begin{align}\label{equ:se}
\begin{split}
& SE(\mathbf{y}^{\ast}|\mathbf{x}^{\ast}) = \mathbb{H}(\mathbf{y}^{\ast}|\mathbf{x}^{\ast}) \\
&=-p(veh|\mathbf{x}^{\ast}) \log p(veh|\mathbf{x}^{\ast})-p(\neg veh|\mathbf{x}^{\ast}) \log p(\neg veh|\mathbf{x}^{\ast}) \\
&\approx-\frac{1}{N}\sum_{i=1}^{N} s_{\mathbf{x}^{\ast}}^{i} \log \frac{1}{N}\sum_{i=1}^{N} s_{\mathbf{x}^{\ast}}^{i}- (1-\frac{1}{N}\sum_{i=1}^{N} s_{\mathbf{x}^{\ast}}^{i}) \log (1-\frac{1}{N}\sum_{i=1}^{N} s_{\mathbf{x}^{\ast}}^{i}). 
\end{split}
\end{align}
SE captures the uncertainty in the prediction output $p(veh|\mathbf{x}^{\ast})$. In this work, we use the natural logarithm, and the SE score ranges in $[0,ln(2)]$. When $p(veh|\mathbf{x}^{\ast})=0$ or $1$, the network is certain with its prediction, resulting SE$=0$. The SE score reaches its peak when the network is most uncertain with its prediction, namely, $p(veh|\mathbf{x}^{\ast})=\frac{1}{2}$.

Different from SE, MI calculates the models' confidence in the output. It measures the information difference between the prediction probability $p(veh|\mathbf{x}^{\ast})$ and the posterior of model parameters \cite{Gal2016Uncertainty} according to

\begin{flalign}\label{equ:mi}
\begin{split}
&MI(\mathbf{y}^{\ast},\mathbf{W}) = \mathbb{H}(\mathbf{y}^{\ast}|\mathbf{x}^{\ast})- \mathbb{E}_{p(\mathbf{W}|\mathbf{X},\mathbf{Y})} \mathbb{H}(\mathbf{y}^{\ast}|\mathbf{x}^{\ast},\mathbf{W})   \\ & = SE(\mathbf{y}^{\ast}|\mathbf{x}^{\ast}) +
\mathbb{E}_{p(\mathbf{W}|\mathbf{X},\mathbf{Y})}\bigg( p(veh|\mathbf{x}^{\ast},\mathbf{W}) \log p(veh|\mathbf{x}^{\ast},\mathbf{W}) \\ & + p(\neg veh|\mathbf{x}^{\ast},\mathbf{W})) \log p(\neg veh|\mathbf{x}^{\ast},\mathbf{W})\bigg) \\
& \approx -\frac{1}{N}\sum_{i=1}^{N} s_{\mathbf{x}^{\ast}}^{i} \log \frac{1}{N}\sum_{i=1}^{N} s_{\mathbf{x}^{\ast}}^{i} \\ & + \frac{1}{N}\sum_{i=1}^{N}\bigg( s_{\mathbf{x}^{\ast}}^{i} \log s_{\mathbf{x}^{\ast}}^{i} + (1-s_{\mathbf{x}^{\ast}}^{i}) \log (1-s_{\mathbf{x}^{\ast}}^{i}) \bigg).
\end{split}
\end{flalign}

MI ranges between $[0,1]$, with a large value indicating high epistemic classification uncertainty. 

\subsubsection{Extracting 3D Bounding Box and Epistemic Spatial Uncertainty}
We estimate the bounding box position of a region proposal in the Lidar coordinate frame, denoted as $\mathbf{l}_{\mathbf{x}^{\ast}}$. For this purpose, we calculate the mean value of the regression outputs of $N$ forward passes. To do this, we first transform the bounding box prediction into the Lidar coordinate frame $\mathbf{\hat{v}}_{\mathbf{x}^{\ast}} = h\mathbf{v}_{\mathbf{x}^{\ast}} + \hat{\mathbf{v}}_{0_{\mathbf{x}^{\ast}}}$, the mean is calculated accordingly
\begin{equation}\label{equ:regression}
\mathbf{l}_{\mathbf{x}^{\ast}} \approx \frac{1}{N}\sum_{i=1}^{N} \hat{\mathbf{v}}_{\mathbf{x}^{\ast}}^{i}.
\end{equation}
To estimate the epistemic spatial (regression) uncertainty in the bounding box prediction, we use the total variance of the covariance matrix of $N$ forward pass regressions. The covariance matrix is calculated via $C(\mathbf{x}^{\ast}) = \frac{1}{N}\sum_{i=1}^{N} \mathbf{\hat{v}}_{\mathbf{x}^{\ast}}^{i} \mathbf{\hat{v}}_{\mathbf{x}^{\ast}}^{i^T} - \mathbf{l}_{\mathbf{x}^{\ast}} \mathbf{l}_{\mathbf{x}^{\ast}}^{T}$. 
The total variance $TV(\mathbf{x}^{\ast})$ is the trace of the covariance matrix according to
\begin{equation}\label{equ:total_variance}
TV(\mathbf{x}^{\ast}) = trace\Big(C(\mathbf{x}^{\ast})\Big).
\end{equation}

This score ranges in $[0,+\infty)$, with a large value indicating high epistemic spatial uncertainty.

\subsection{Capturing Aleatoric Uncertainty}\label{subsec:aleatoric}
So far, we have explained how to use the dropout sampling technique to approximate the posterior distribution and to extract epistemic uncertainty. To capture the aleatoric uncertainty, we need to model the distribution $p(\mathbf{y}^{\ast}|g^{\mathbf{W}}(\mathbf{x}^{\ast}))$. In the classification task we can model it by the softmax function, i.e. $p(\mathbf{y}^{\ast}|g^{\mathbf{W}}(\mathbf{x}^{\ast}))=softmax(g^{\mathbf{W}}(\mathbf{x}^{\ast})) = s_{\mathbf{x}^{\ast}}$, with $\mathbf{y}^{\ast}$ referring to the object labels. In the 3D bounding box regression task, we model the observation likelihood as a multi-variate Gaussian distribution with diagonal covariance matrix via
\begin{equation}\label{equ:alea}
\begin{split}
&p(\mathbf{y}^{\ast}|g^{\mathbf{W}}(\mathbf{x}^{\ast}))=\mathcal{N}(g^{\mathbf{W}}(\mathbf{x}^{\ast}),\Sigma(\mathbf{x}^{\ast})), \\
&\Sigma(\mathbf{x}^{\ast})=diag(\sigma^2_{\mathbf{x}^{\ast}}).
\end{split}
\end{equation}

Here, $\mathbf{y}^{\ast}$ is the bounding box prediction. The parameter $\sigma^2_{\mathbf{x}^{\ast}}$ is a $24$-dimensional vector, with each element representing an observation noise of a prediction in the bounding box regression $\mathbf{v}_{\mathbf{x}^{\ast}}$. We can obtain the observation noises by adding an output regression layer in our vehicle detection network. In addition to the softmax scores and bounding box regressions, now it allows us to predict the observation noises $\lambda_{\mathbf{x}^{\ast}}$, where we use $\lambda_{\mathbf{x}^{\ast}} \coloneqq \log(\sigma^2_{\mathbf{x}^{\ast}})$ for numerical stability. In the training phase, we modify the cost function for the bounding box regression through 
\begin{equation}\label{equ:new_cost_func}
L_{out\text{-}reg}= \frac{1}{2} \exp(-\lambda^T) \Vert \mathbf{v}_{gt} - \mathbf{v} \Vert + \frac{1}{2} \lambda^T\mathbf{1}.
\end{equation}
As \cite{kendall2017uncertainties} mentioned, this loss function can increase the network robustness when learning from noisy dataset. When the training data has high aleatoric uncertainty, the model is penalized by the $\frac{1}{2} \lambda^T\mathbf{1}$ term and ignores the residual term $\Vert \mathbf{v}_{gt} - \mathbf{v} \Vert$ since $\exp(-\lambda^T)$ becomes small. Consequently, the data with high uncertainty contributes less to the loss. Note that instead of performing approximate inference when extracting epistemic uncertainty, here we perform MAP inference.
\subsection{Implementation}
Training the neural network is achieved in a multi-loss end-to-end learning fashion. We use smooth $l_1$ loss and cross-entropy loss for the oriented 3D box regression ($L_{out\text{-}reg}$) and classification ($L_{out\text{-}cls}$) respectively. We also use the same loss function for region proposals and object category outputs in the Region Proposal Network ($L_{rpn\text{-}cls}$ and $L_{rpn\text{-}reg}$). The final loss function is formulated as:
\begin{equation}
Loss = \gamma_1L_{rpn\text{-}cls} + \gamma_2L_{rpn\text{-}reg} + \gamma_3 L_{out\text{-}cls} + \gamma_4L_{out\text{-}reg},
\end{equation}
where we set $\gamma_1,\gamma_3=1$ and $\gamma_2,\gamma_4=0.05$.
We employ $L_2$ regularization and dropout with dropout rate $p=0.5$ to prevent over-fitting and to perform posterior distribution approximation. The network is trained $70000$ steps using Adam optimizer and a learning rate of $0.0001$. We then reduce the learning rate to $0.00001$ and train another $10000$ steps for fine tuning.

To solely extract epistemic uncertainty, the network performs $N$ number of forward passes with dropout during the test time. We fix the dropout rate at $p=0.5$, though it can be tuned by grid-searching to maximize the validation log likelihood \cite{Gal2016Uncertainty} or by using gradient methods \cite{gal2017concrete}. A dropout rate of $0.5$ introduces the highest epistemic uncertainty in the model. We also fix $N=40$. To solely extract the aleatoric uncertainty, the network is trained with the loss function modified according to Eq.~\ref{equ:new_cost_func}. It performs the feed-forward pass only once without dropout during the test time. To extract both epistemic and aleatoric uncertainty, the network trained with Eq.~\ref{equ:new_cost_func} needs to perform multiple feed-forward passes with dropout. 

\section{\textbf{Experimental Results}}
\label{sec:results}
In this section, we experimentally evaluate our proposed probabilistic Lidar 3D object detection network. We show that modeling the aleatoric uncertainty improves the vehicle detection performance. We also show that the network captures model uncertainty and observation uncertainty, which behave differently from each other: The model (epistemic) uncertainty is influenced by the detection accuracy, but is unaffected by the vehicle distance. Conversely, the observation (aleatoric) uncertainty is associated with the vehicle distance and occlusion, and has little relationship to detection accuracy. 
\subsection{Experimental Setup}
Our proposed method was evaluated on the KITTI raw dataset \cite{Geiger2013IJRR}. Among the dataset, we randomly chose $29$ drives for training, and another $6$ drives for testing (drive $0001$, $0015$, $0039$, $0052$, $0070$, $0086$). This corresponds to $9918$ training frames and $2010$ testing frames. We considered the Lidar point clouds in the ranges $x=0-100$ m, $y=\-30-30$ m, $z=\-3.5-0.6$ m (The lidar coordinate system is illustrated in Fig.~\ref{fig:introduction}). With a discretization resolution of $0.1$ m, each bird's eye view input channel was a 2D image with $1000\times600$ pixels. We only evaluated the detections that were visible in the front-view of the camera.

The 3D vehicle proposals were considered to be detected when their probability scores were larger than $0.5$. Their performance was evaluated using Intersection Over Union (IoU) threshold ranging from $0.1$ to $0.8$. The IoU scores rate the similarity between a predicted vehicle and its ground truth. A higher IoU value with the ground truth indicates a more accurate prediction. 
\subsection{3D Vehicle Detection Performance}
We first evaluated the 3D vehicle detection performance of our proposed network that captures epistemic and aleatoric uncertainty. As a baseline method, we trained a vehicle detector called \textit{Non-Bayesian} which did not explicitly model the observation noise $P(\mathbf{v}|g^{\mathbf{W}}(\mathbf{x}))$ and the weight posterior distribution $P(\mathbf{W}|\mathbf{X},\mathbf{Y})$. We then compared the baseline method with the vehicle detectors that captured epistemic uncertainty (\textit{Epistemic}), aleatoric uncertainty (\textit{Aleatoric}), or both (\textit{Epistemic+Aleatoric}). We used $F_1$ scores ($F_1 = \frac{2PR}{P+R}$) at different IoU threshold to evaluate the detection performance, where $P$ indicates the 3D bounding box precision, and $R$ the recall value. Results are illustrated in Tab.~\ref{tab:pr_curve}. The vehicle detectors that captured aleatoric uncertainty (\textit{Aleatoric} and \textit{Epistemic+Aleatoric}) consistently outperformed the baseline method, with an improvement of $F_1$ scores by $1\%-5\%$. This is because modeling the aleatoric uncertainty has increased the model robustness when dealing with noisy input data. The network \textit{Epistemic}, conversely, slightly underperformed the baseline method, as network capability was reduced by dropping out some of its hidden units. 

\begin{table}[htbp]
\caption{$F_1$ score comparison for different vehicle detectors}
\begin{center}
\scalebox{0.72}{
\begin{tabular}{c|c|c|c|c|c|c|c|c}
& \multicolumn{8}{c}{\textbf{IoU threshold}} \\
\hline
\textbf{Network} & \textbf{0.1}& \textbf{0.2} & \textbf{0.3} & \textbf{0.4} & \textbf{0.5} & \textbf{0.6} & \textbf{0.7} & \textbf{0.8}\\
\hline
\hline
Non-Bayesian (baseline) & $0.737$ & $0.724$ & $0.697$ & $0.675$ & $0.634$ & $0.508$ & $0.247$ & $0.043$ \\
\hline
Epistemic & $0.732$ & $0.717$ & $0.691$ & $0.668$ & $0.625$ & $0.498$ & $0.245$ & $0.042$ \\
\hline
Aleatoric & \textbf{0.750} & \textbf{0.736} & \textbf{0.716} & \textbf{0.688} & \textbf{0.651} & \textbf{0.533} & \textbf{0.253} & \textbf{0.044} \\
\hline
Epistemic+Aleatoric & $0.745$ & $0.730$ & $0.704$ & $0.679$ & $0.636$ & $0.517$ & $0.248$ & $0.043$ \\
\hline
\end{tabular}}
\label{tab:pr_curve}
\end{center}
\end{table}

\subsection{Understanding the Epistemic Uncertainty in 3D Object Detection} \label{subsec:exp_epistemic}

\begin{figure*}[htpb]
    \centering
    \begin{minipage}{0.96\textwidth}
	\centering
	\subfigure[Averaged Shannon entropy for epistemic uncertainty]{\label{fig:se_mean}\includegraphics[width=0.32\textwidth]{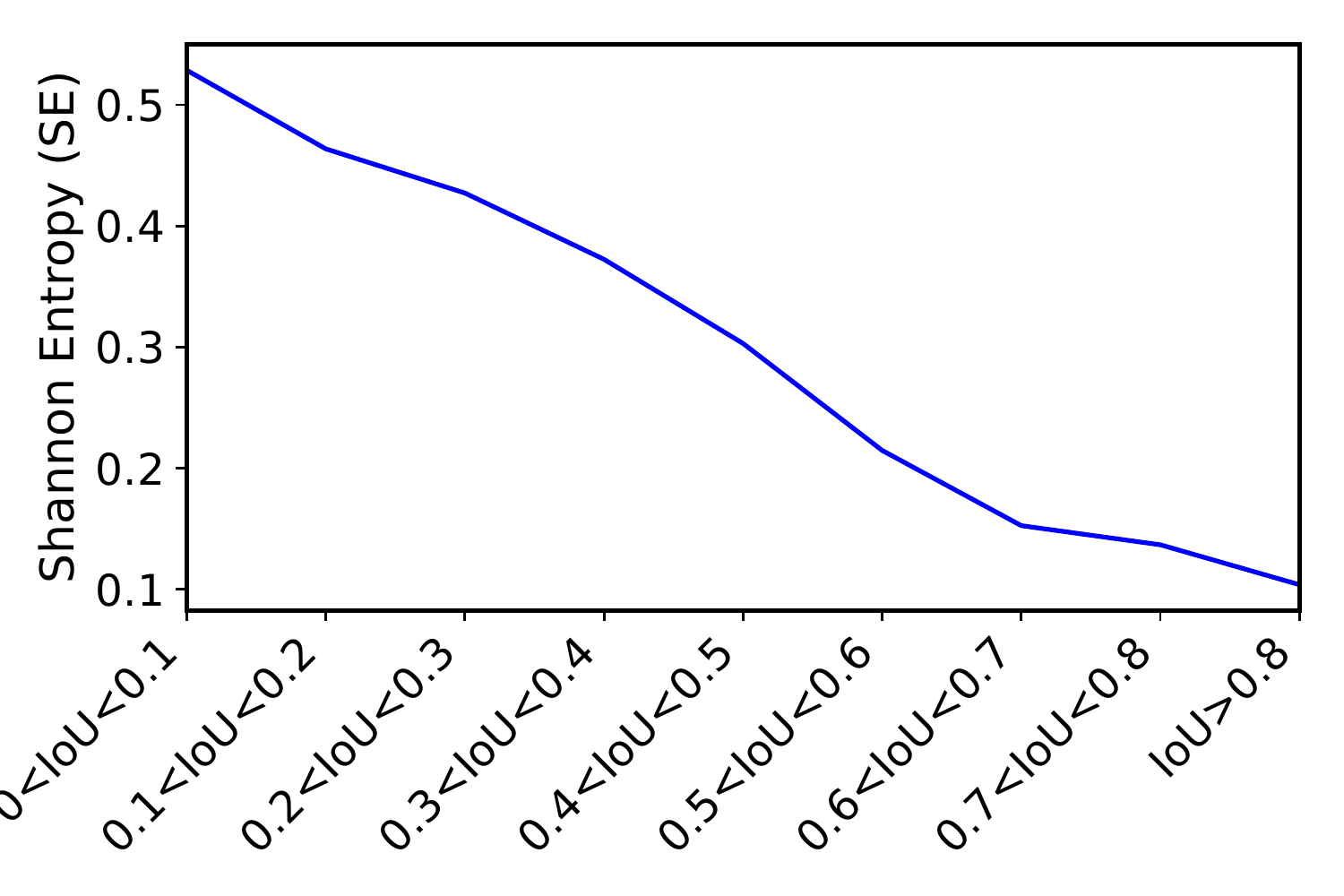}}
	\subfigure[Averaged mutual information for epistemic uncertainty]{\label{fig:mi_mean}\includegraphics[width=0.32\textwidth]{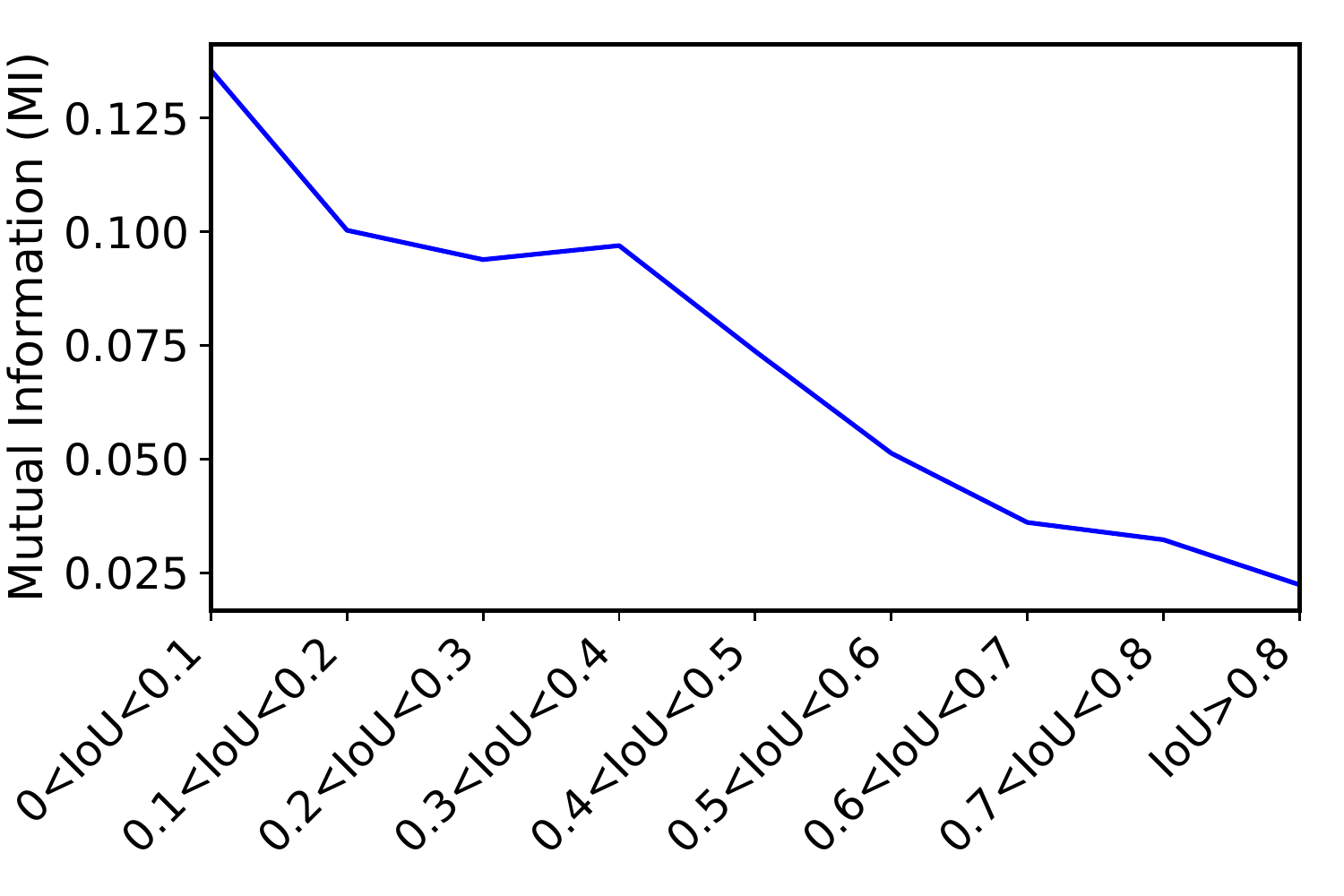}}
	\subfigure[Averaged total variance for epistemic uncertainty]{\label{fig:regression_mean}\includegraphics[width=0.32\textwidth]{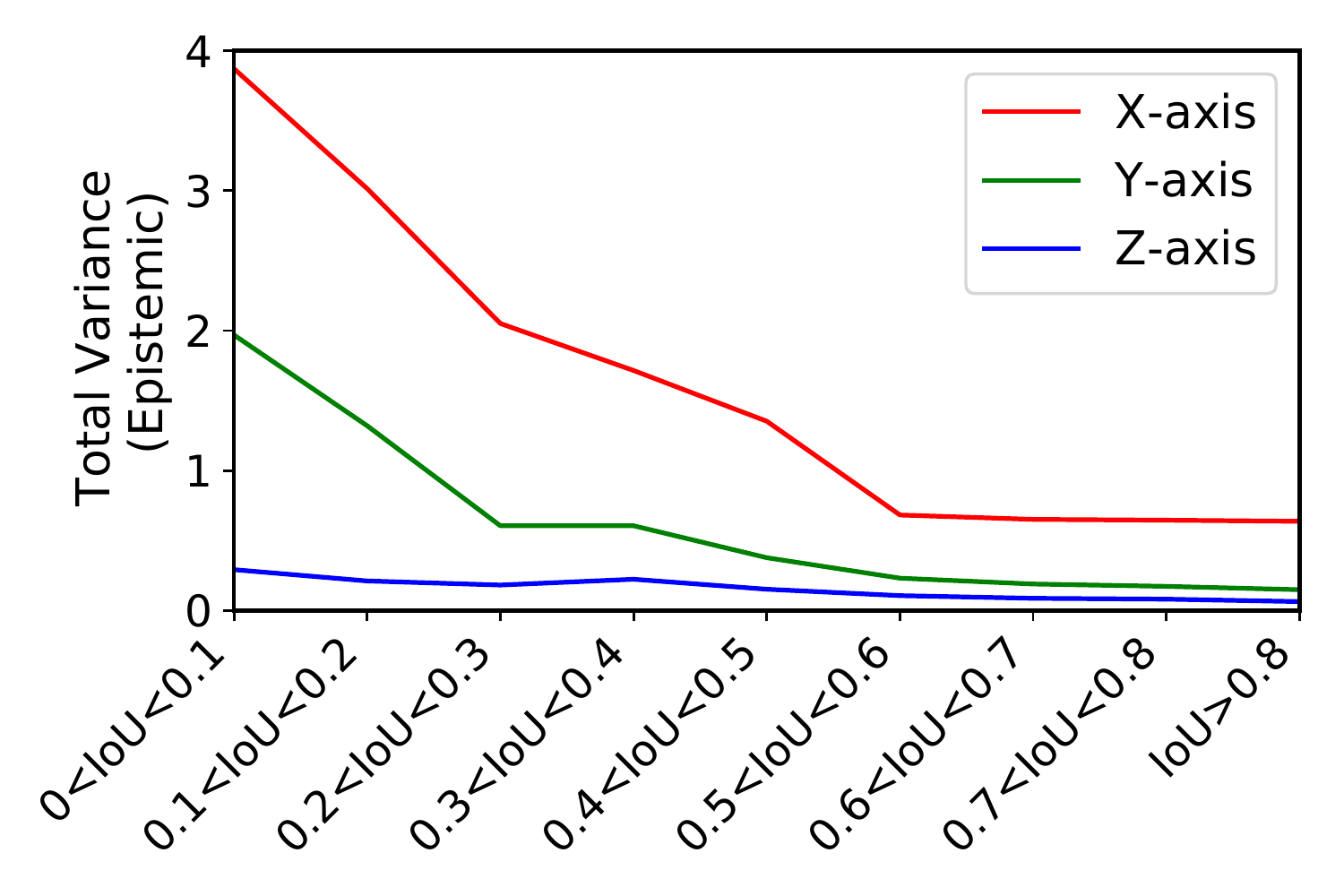}}
	\caption{Averaged espistemic uncertainty of predicted samples at differnet IoU intervals. The horizontal axis represents the increasing IoU values, and the vertical axis represents the uncertainty measurement. (a), (b): Shannon Entropy and Mutual Information estimates for epistemic classification uncertainty. (c): Total Variance for epistemic spatial uncertainty. The total variance in $x$, $y$, and $z$ axes were calculated.}
\end{minipage}
\end{figure*}

\begin{figure*}[htpb]
    \centering
    \begin{minipage}{0.96\textwidth}
	\centering
	\subfigure[Epistemic classification uncertainty at IoU$>0.8$]{\label{fig:tp_cls}\includegraphics[width=0.22\textwidth]{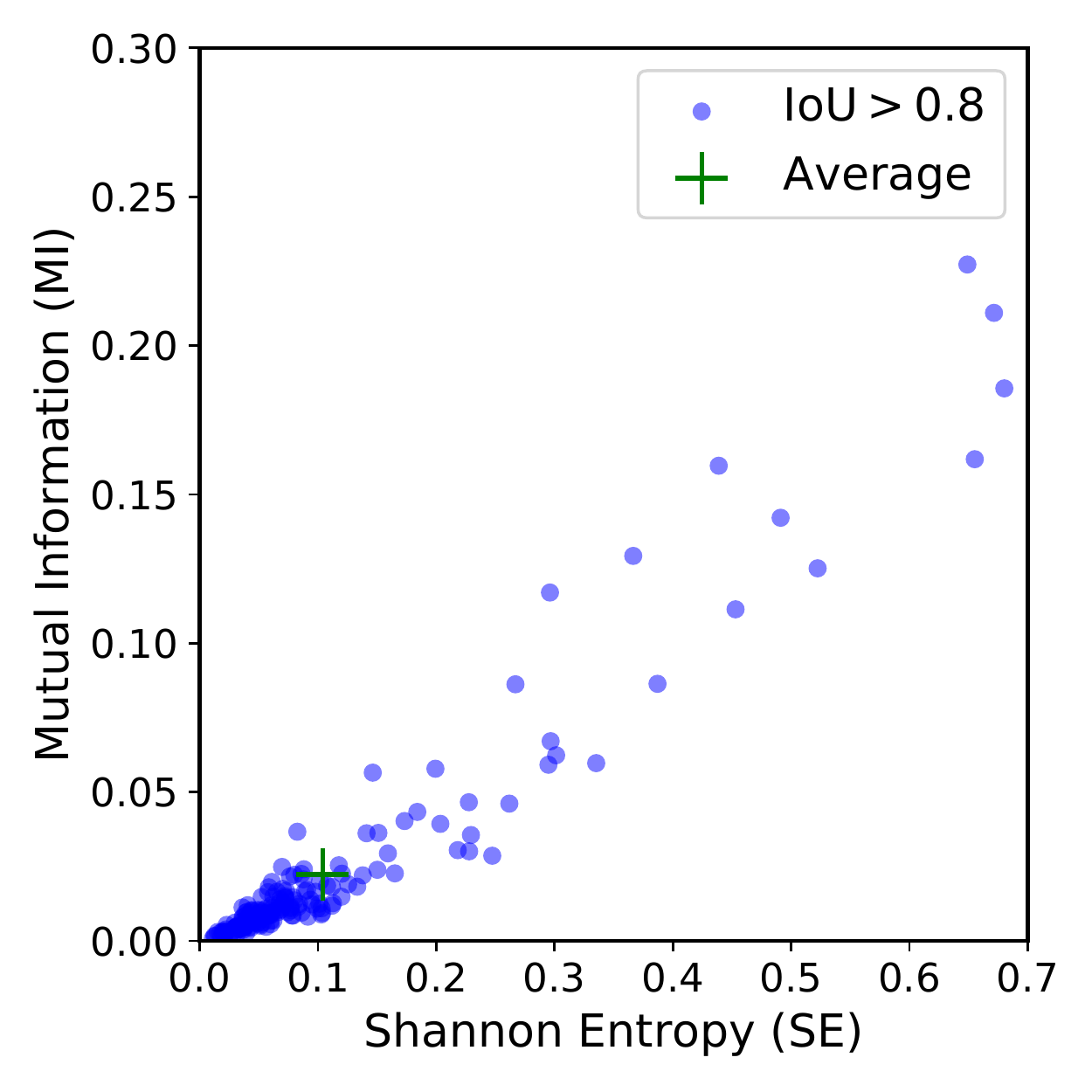}}
	\subfigure[Epistemic classification uncertainty at IoU$<0.1$]{\label{fig:fp_cls}\includegraphics[width=0.22\textwidth]{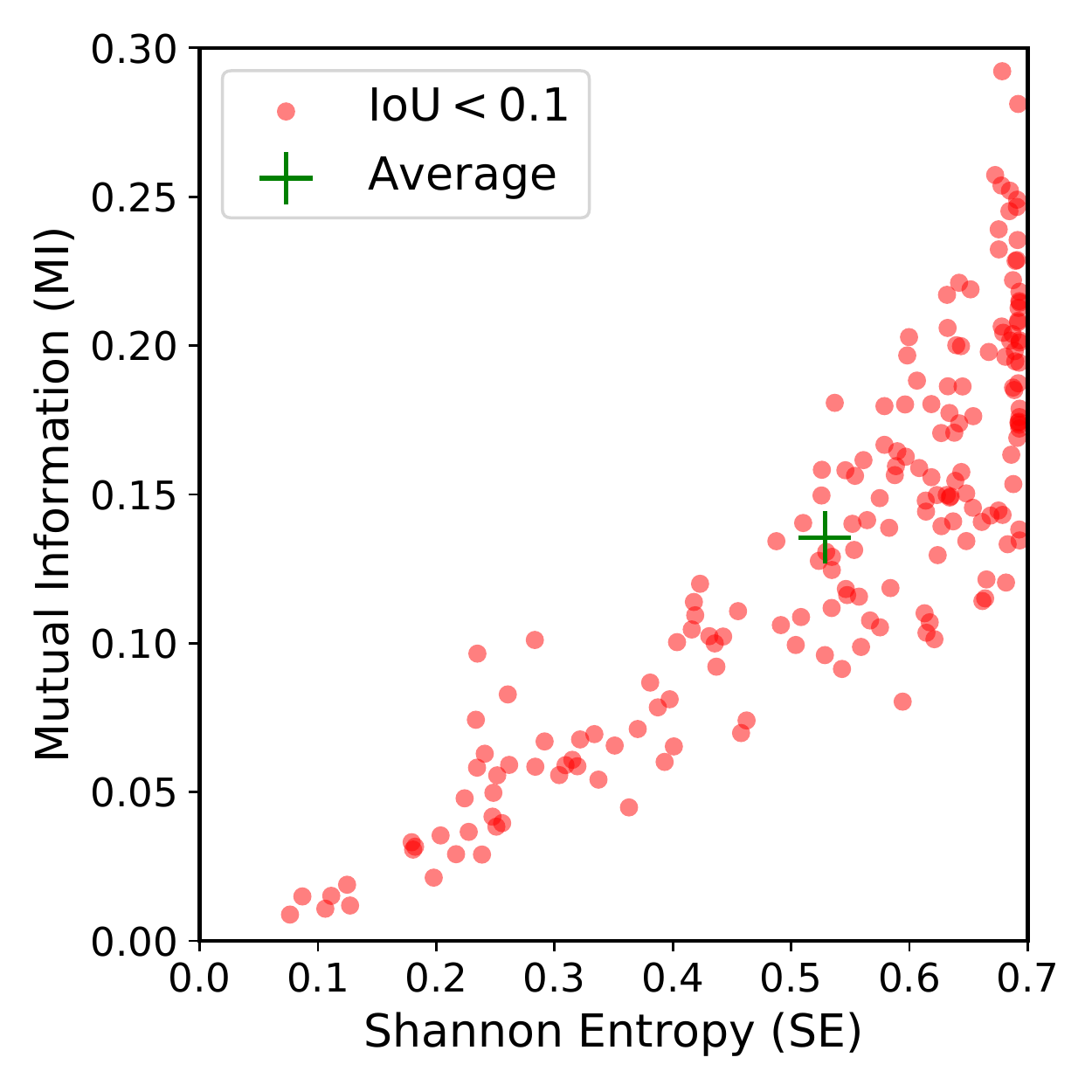}}
	\subfigure[Epistemic spatial uncertainty at IoU$>0.8$]{\label{fig:tp_reg}\includegraphics[width=0.26\textwidth]{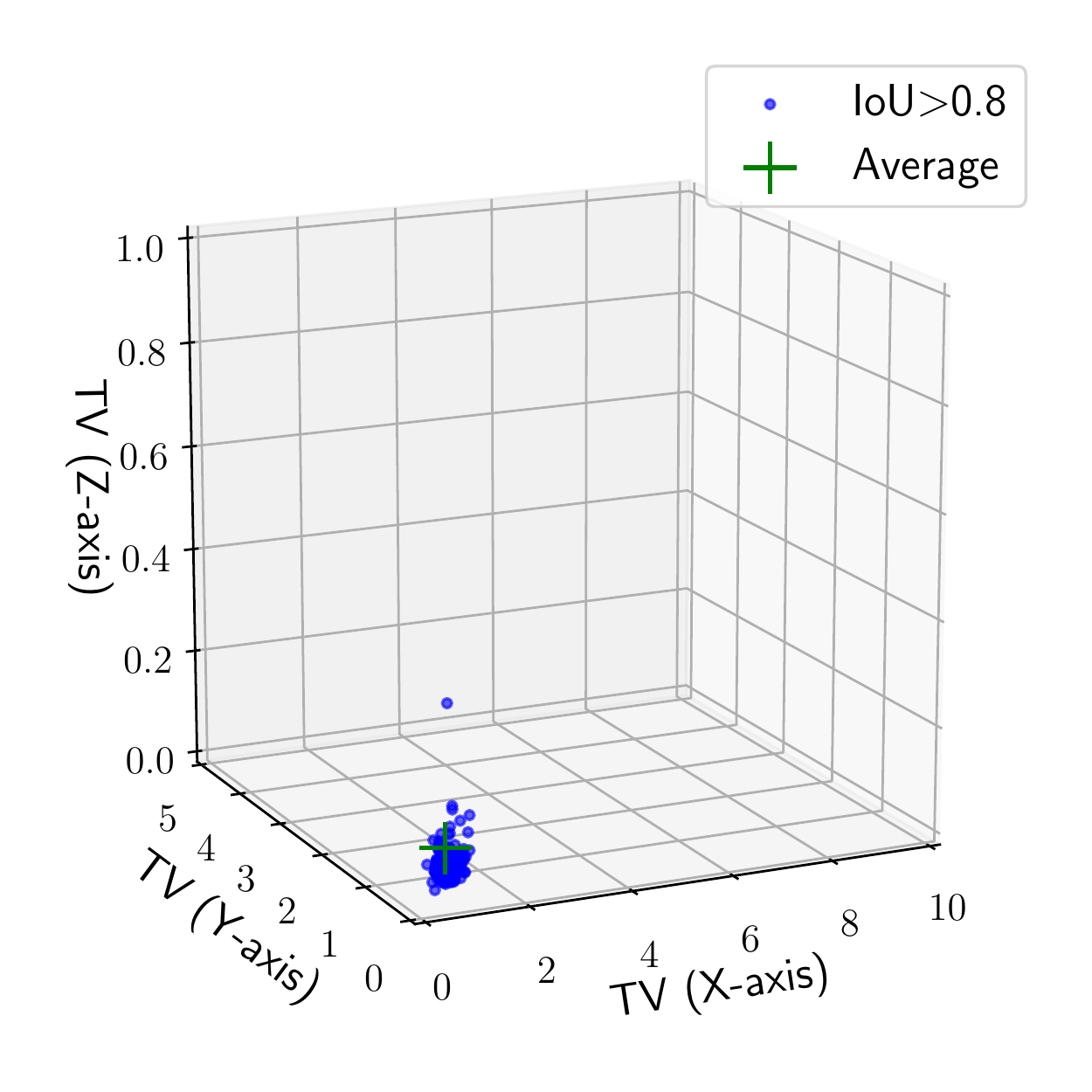}}
    \subfigure[Epistemic spatial uncertainty at IoU$<0.1$]{\label{fig:fp_reg}\includegraphics[width=0.26\textwidth]{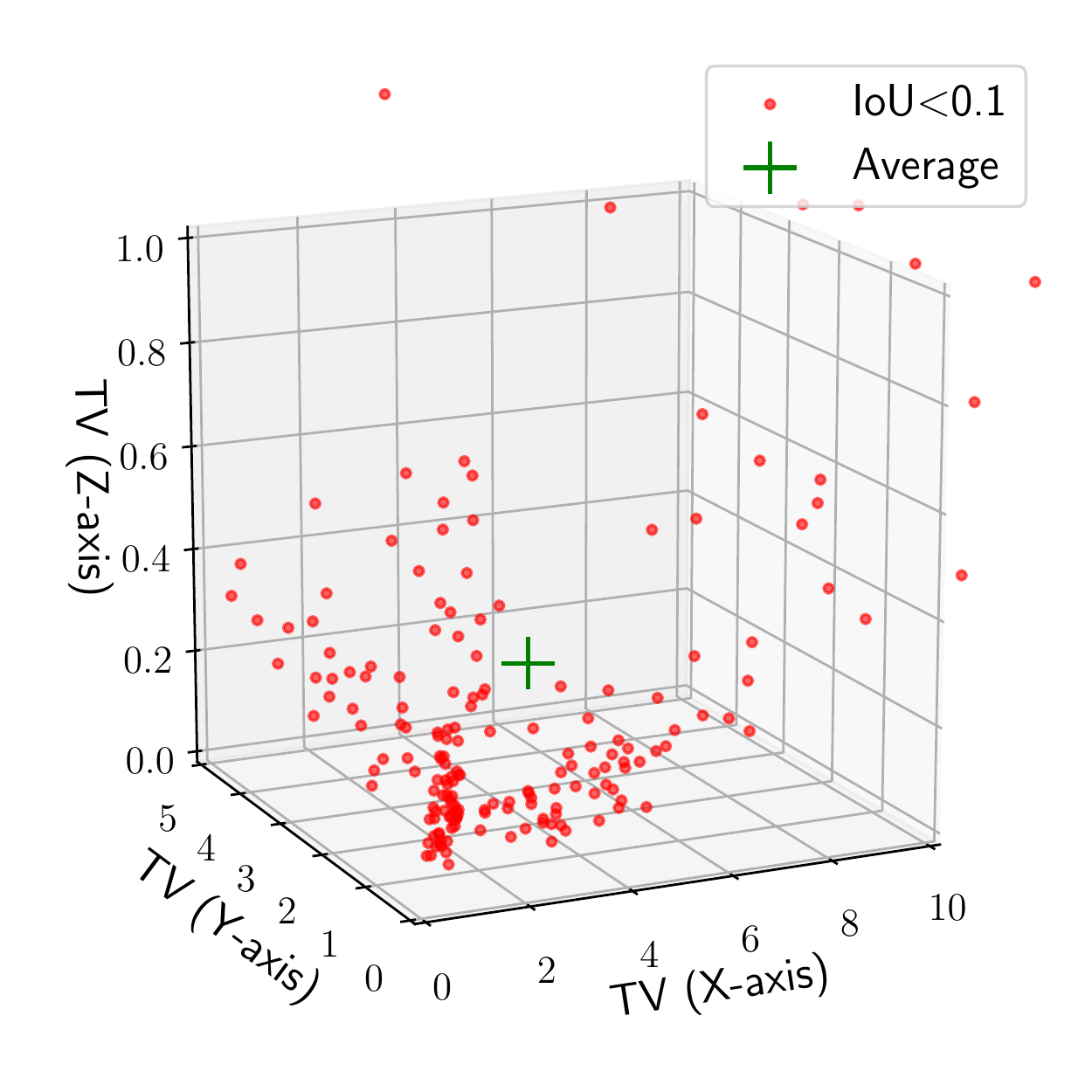}}
	\caption{The epistemic uncertainty estimates for each detection at IoU$<0.1$ and IoU$>0.8$. (a), (b): classification uncertainty; (c), (d): 3D bounding box spatial uncertainty. The total variances in $x$, $y$, and $z$ axes were calculated.}
\end{minipage}
\end{figure*}
We then evaluated how our proposed network captured the epistemic uncertainty in 3D object detection task. Fig.~\ref{fig:se_mean} and Fig.~\ref{fig:mi_mean} demonstrate the mean values of Shannon Entropy (SE) and Mutual Information (MI) averaged over all predicted samples that lay within different IoU intervals. Predictions with higher IoU scores were more accurate. Results show that SE behaved similar to the MI score. Both scores were associated well with the prediction accuracy, as they became smaller with increasing IoU. We also evaluated the epistemic uncertainty in the bounding box regression. Following Eq.~\ref{equ:total_variance}, we calculated the mean values of total variance in $x$, $y$, and $z$ axes. Fig.~\ref{fig:regression_mean} shows that the total variance in the $x\-$axis was the largest, indicating that the network was most uncertain to estimate the vehicles' positions in the $x$ direction. The total variance in the $z\-$axis was the smallest, as there was little variance in vehicles' height. Fig.~\ref{fig:regression_mean} also shows that the regression uncertainty was affected by the prediction accuracy. The total variance decreased for larger IoU values.  

To better understand the epistemic uncertainty, we plot the SE and MI values for each predicted vehicle with IoU $>0.8$, referring to the most accurate detections (Fig.~\ref{fig:tp_cls}), and IoU$<0.1$ $-$ the worst detections (Fig.~\ref{fig:fp_cls}). The SE and MI values for detections with IoU $>0.8$ mostly lay on the bottom left of the figure, showing the low epistemic classification uncertainty. In contrast, detections with IoU $<0.1$ were widely spread in Fig.~\ref{fig:fp_cls}, demonstrating higher uncertainty. We also analyzed the distribution of total variance for IoU $>0.8$ and IoU $<0.1$. Fig.~\ref{fig:tp_reg}, Fig.~\ref{fig:fp_reg} show large epistemic regression uncertainties, when the network made inaccurate detections. In conclusion, our proposed Lidar 3D object detector captured reliable epistemic classification and regression uncertainty. The objects with low classification and regression uncertainty were more likely to be detected accurately.

\begin{figure}[htbp]
    \centering
	\centering
	\subfigure[]{\label{fig:case_study2}\includegraphics[width=0.4\linewidth]{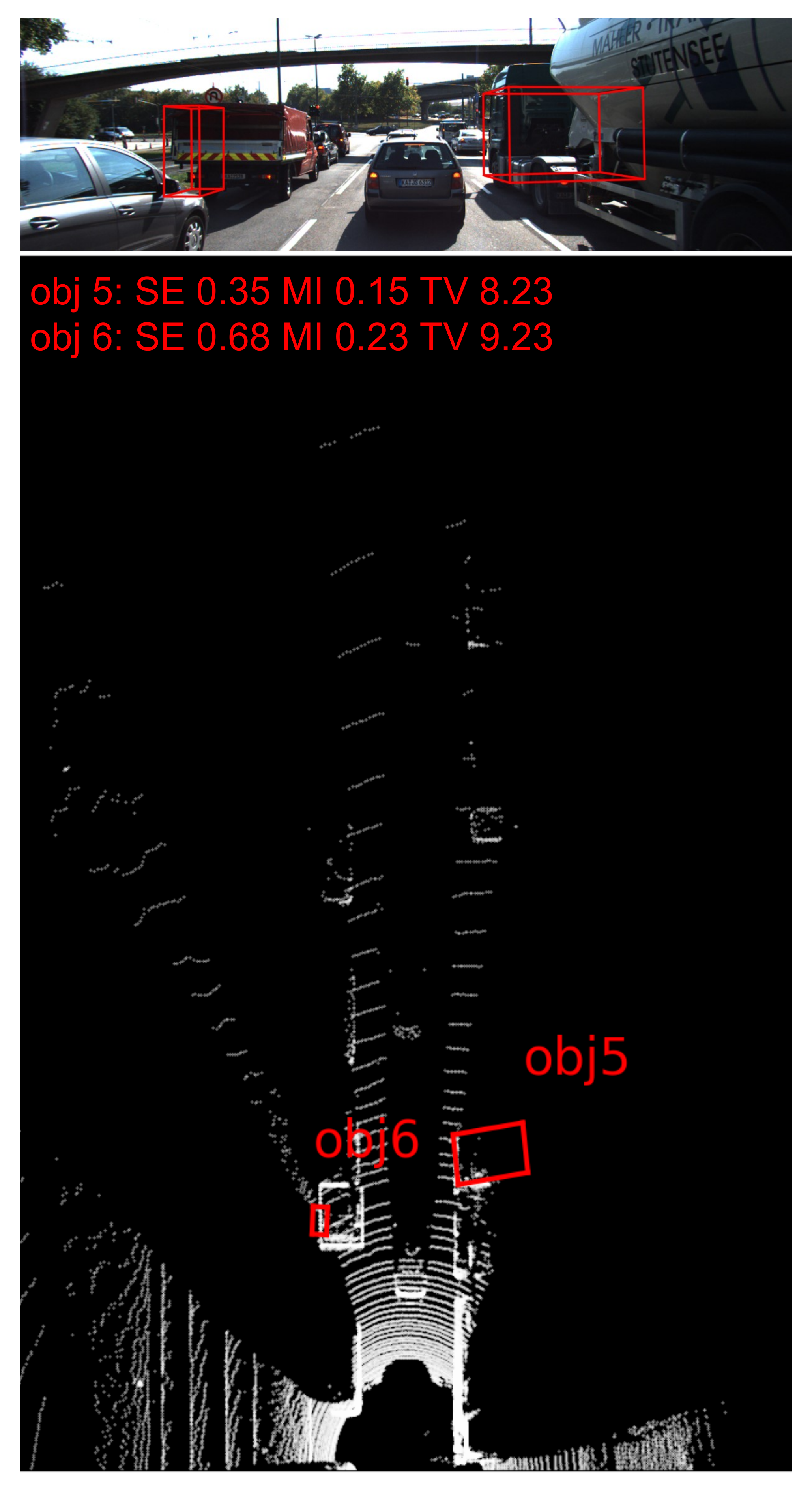}}
	\subfigure[]{\label{fig:case_study3}\includegraphics[width=0.4\linewidth]{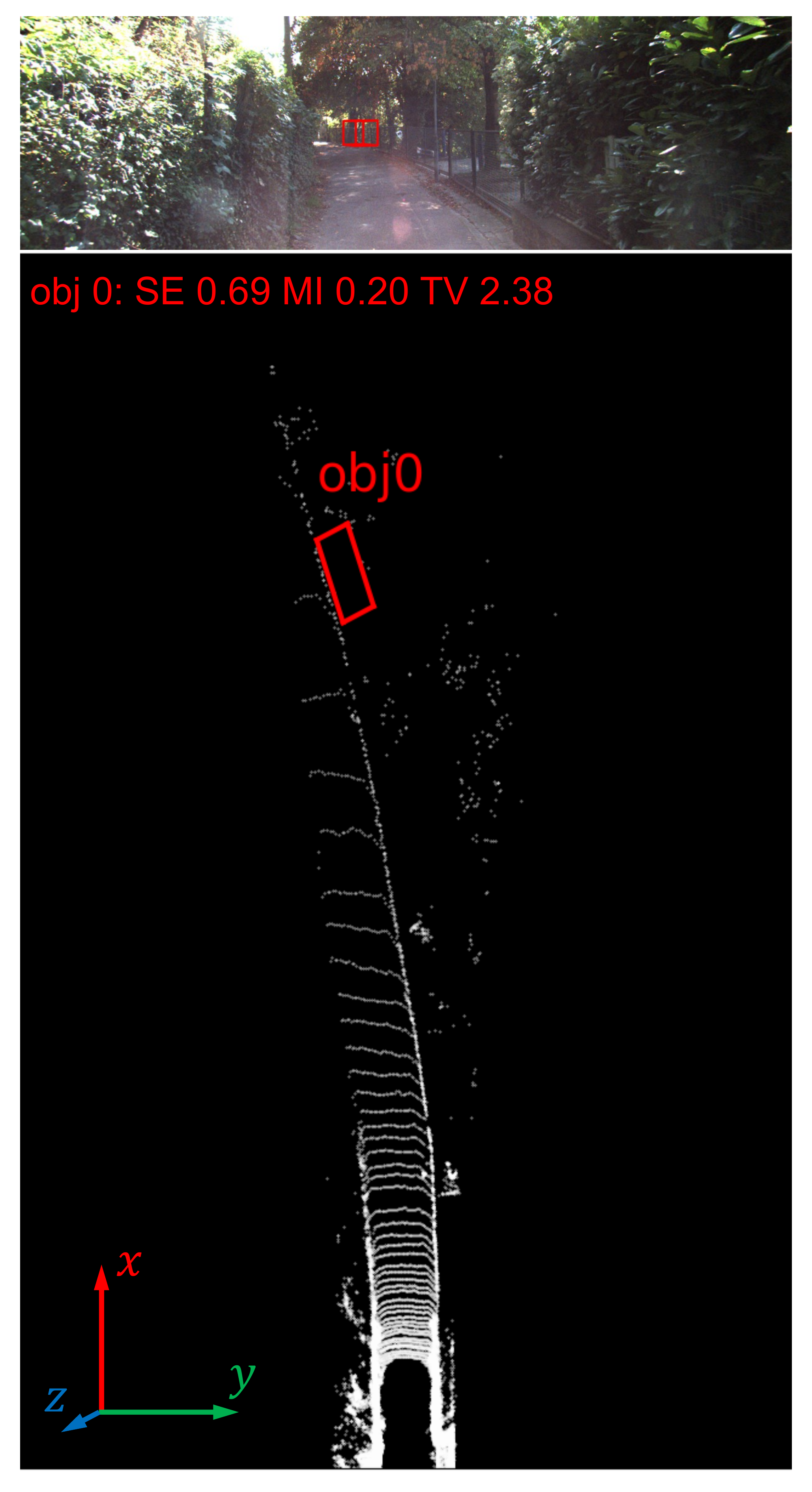}}
    \caption{Illustration of detections with high epistemic uncertainty: (a) big vehicles such as trucks; (b) ghost objects. A larger value indicates higher uncertainty. The camera images are used only for visualization purpose.}
\end{figure}

\textit{Qualitative Observations.} We observed that big vehicles such as vans and trucks as well as the ``ghost" objects (i.e. False Positives) often showed high levels of epistemic classification uncertainty (e.g. obj 5 in Fig.~\ref{fig:case_study2} and obj 0 in Fig.~\ref{fig:case_study3}). Besides, detections with abnormal bounding boxes, e.g. boxes with unusual small length and large height (e.g. obj 6 in Fig.~\ref{fig:case_study2}), often showed high epistemic spatial uncertainty. This is because that these detections differed from our training dataset, which contained no ghost objects, no objects with abnormal shapes, and a few big vehicles. Therefore, our Lidar 3D object detector was uncertain with them, and showed high epistemic uncertainty scores. Such information can be used to efficiently improve the vehicle detector in an active learning paradigm: the detector actively queries the unseen samples with high epistemic uncertainty. More detection results can be found in the supplementary video. 

\subsection{Understanding the Aleatoric Uncertainty in 3D Object Detection}
Finally, we evaluated the aleatoric uncertainty, i.e. the uncertainty that captures the observation noises in Lidar point clouds. Here, we focused on the 3D bounding box regression whose uncertainty is quantified by $\sigma^2$ (Sec.~\ref{subsec:aleatoric}). In this work, we did not explicitly model the parameters to quantify the aleatoric classification uncertainty. We leave it as an interesting topic for the future research.

\begin{figure}[hpbt]
	\centering
		\centering
\includegraphics[width=0.72\linewidth]{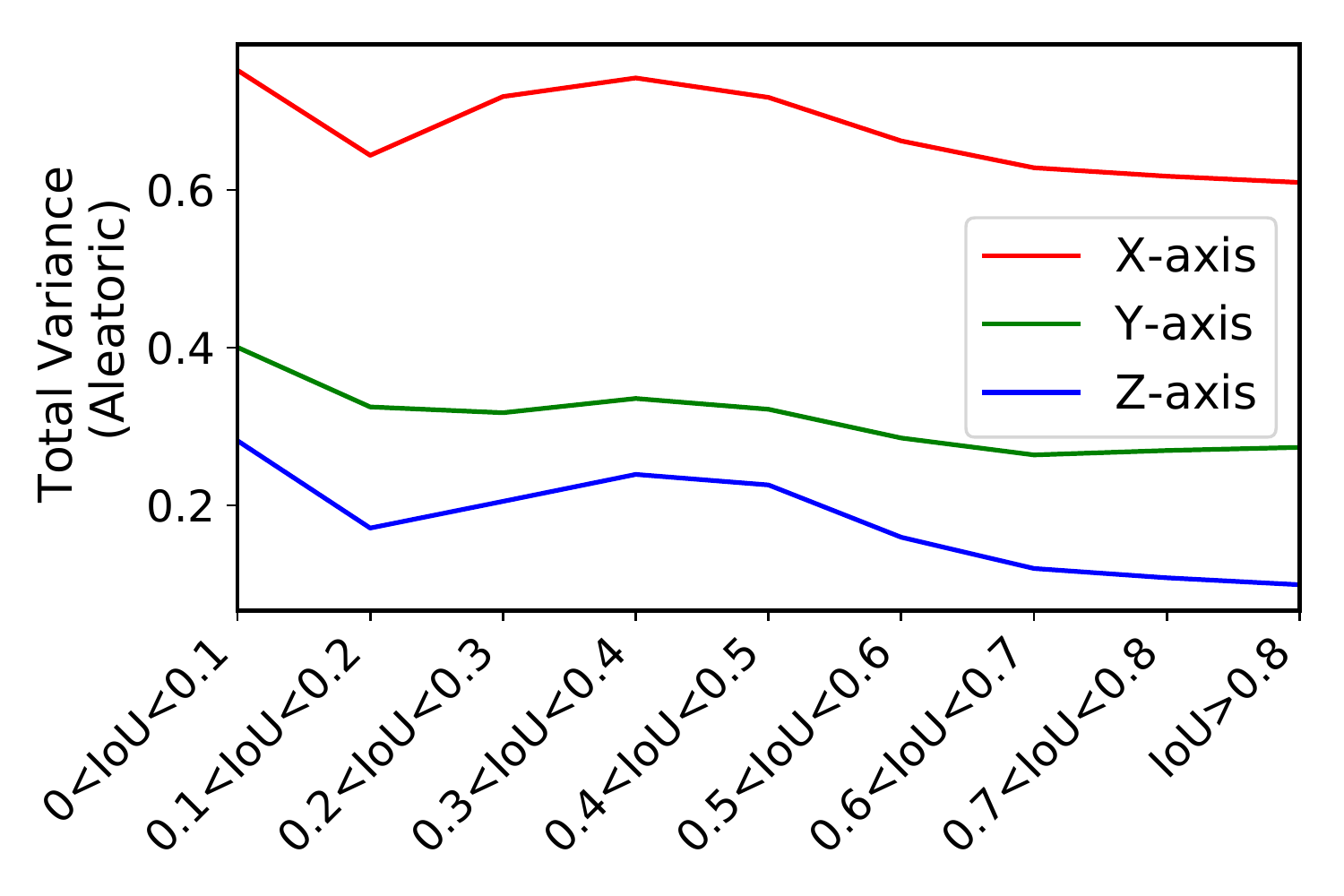}
	\caption{Averaged total variance for aleatoric spatial uncertainty of predicted samples at differnet IoU intervals. The total variance in $x$, $y$, and $z$ axes were calculated.}\label{fig:reg_mean_aleatoric}
\end{figure}

We demonstrated the mean value of total variance of aleatoric spatial uncertainty with regard to different IoU intervals, similar to the experiment in Sec.~\ref{subsec:exp_epistemic}. To do this, we summed up the variance $\sigma^2$ of observation noises of an object in $x$, $y$, $z$ axis, respectively. Fig.~\ref{fig:reg_mean_aleatoric} shows that the aleatoric spatial uncertainty was little related to the detection accuracy, which was different from the epistemic uncertainty shown in Fig~\ref{fig:regression_mean}.

\begin{figure}[htbp]
    \centering
	\includegraphics[width=0.99\linewidth]{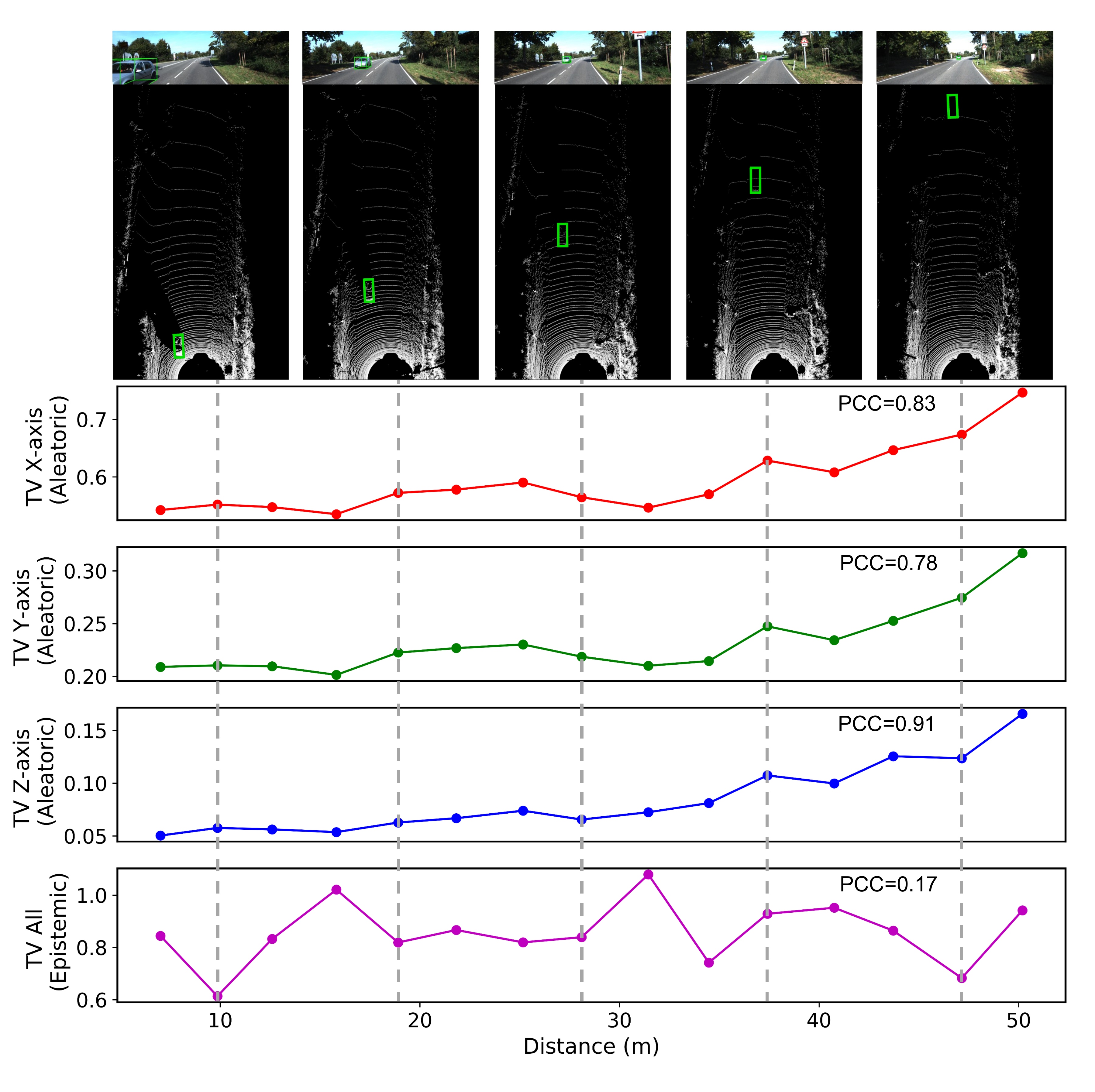}
    \caption{An exemplary evolution of aleatoric and epistemic spatial uncertainty in sequential detections. PCC: Pearson correlation coefficient.}
    \label{fig:dist}
\end{figure}

We further analyzed the behavior of aleatoric uncertainty with regard to the distance between the ego-vehicle and the detected vehicles. We used the Pearson Correlation Coefficient (PCC) to quantify the linear correlation between the distance and the total variance in $x$, $y$, $z$ axes as well as the total variance of the whole covariance matrix, for detections in the test dataset. Tab.~\ref{tab:aleatoric_dist} shows the results. The aleatoric uncertainty was positively correlated with distance, indicating that a more distant vehicle was more difficult to be localized. This is due to the fact that the point clouds of an object become increasingly sparse at a large distance. Contrarily, the epistemic uncertainty showed little relationship with the distance. This conclusion was also supported by an exemplary sequence of vehicle detection shown in Fig.~\ref{fig:dist}. As the car was leaving from the ego-vehicle, the aleatoric uncertainties in $x$, $y$, $z$ axes were continuously increasing, whereas the epistemic uncertainty showed no tendency. 

\begin{table}[htbp]
\begin{center}
\captionsetup{justification=centering}
\caption{Pearson correlation coefficient \\ between distance and spatial uncertainty}
\scalebox{0.96}{
\begin{tabular}{c|c|c|c|c}
\hline
\textbf{Network} & $\bf{X}$ \textbf{axis} & $\bf{Y}$ \textbf{axis} & $\bf{Z}$ \textbf{axis} & \textbf{All} \\
\hline
\hline
Epistemic  & $-0.046$ & $-0.030$ & $0.202$ & $-0.029$ \\
\hline
Aleatoric & \textbf{0.569} & \textbf{0.412} & \textbf{0.497} & \textbf{0.537} \\
\hline
\end{tabular}}
\label{tab:aleatoric_dist}
\end{center}
\end{table}

\begin{figure}[H]
    \centering
	\subfigure[]{\label{fig:case_study0}\includegraphics[width=0.4\linewidth]{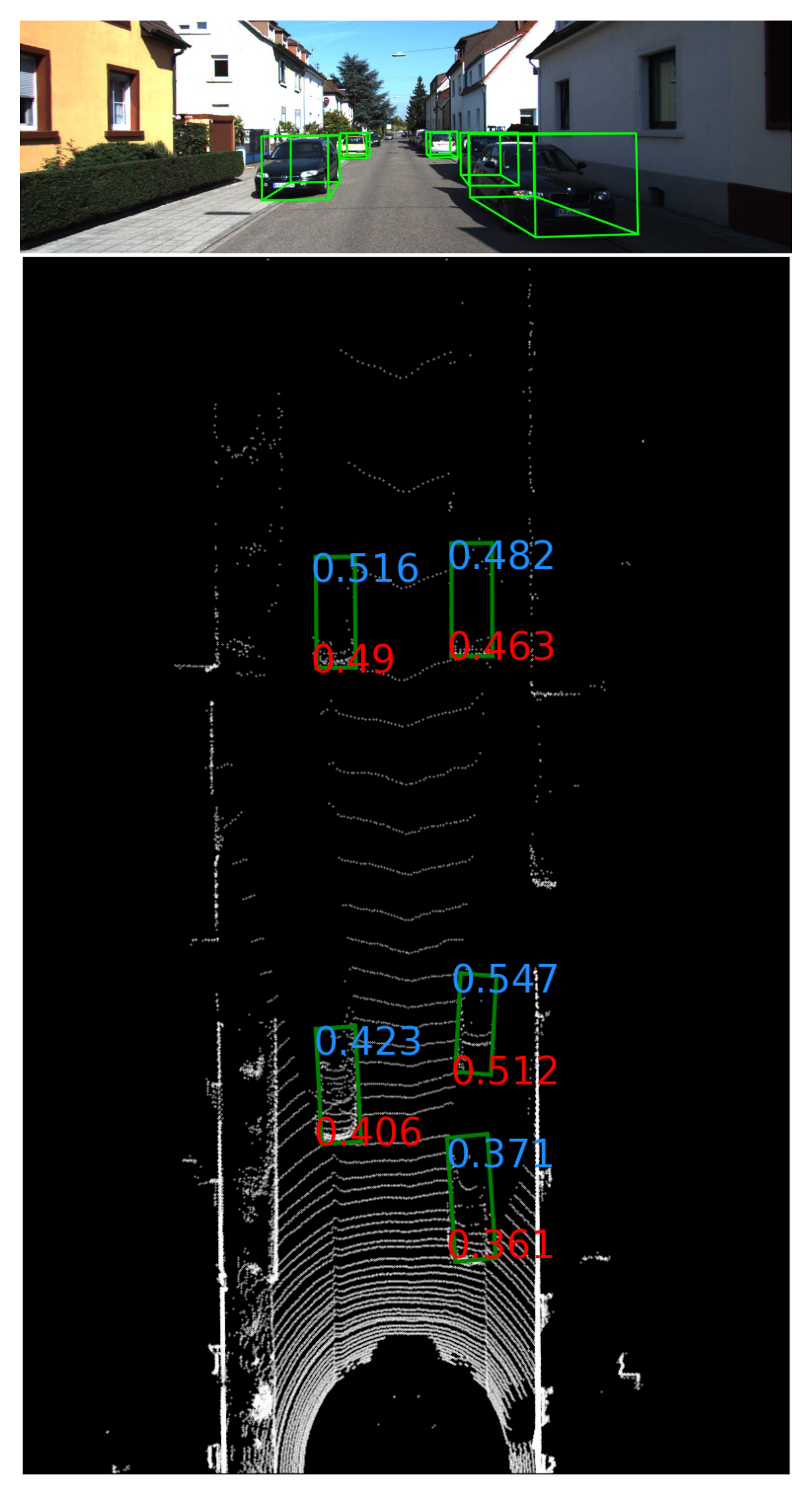}}
	\subfigure[]{\label{fig:case_study1}\includegraphics[width=0.4\linewidth]{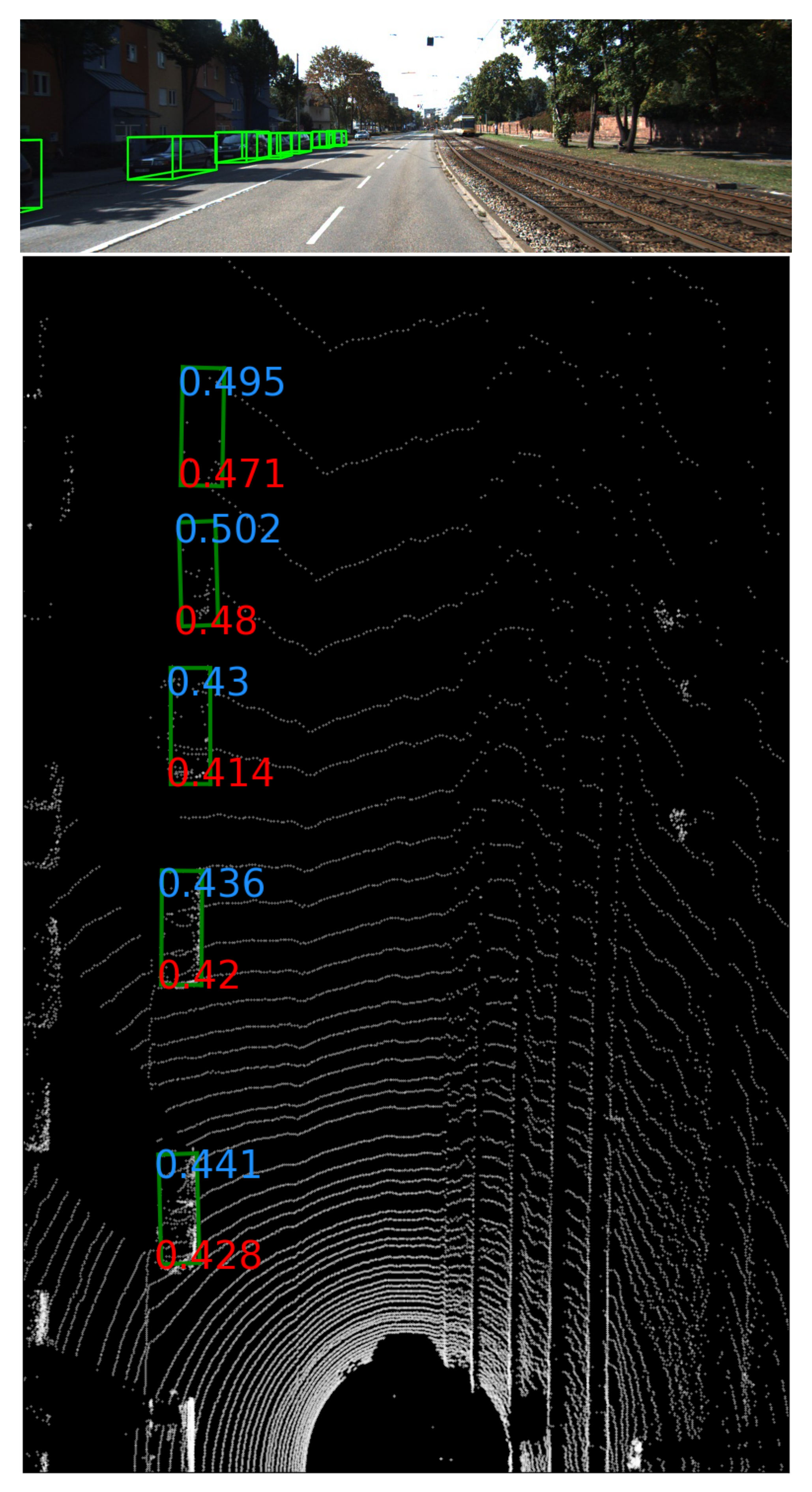}}
    \caption{Illustration of aleatoric spatial uncertainty in the corners of the bounding boxes that face towards to ego-vehicle (marked in red) and occluded corners (marked in blue). A larger value indicates higher uncertainty. }
\end{figure}

The observation noises were not only affected by distance, but by occlusion as well. We found that the corners of the bounding boxes which face directly towards the Lidar sensor consistently display smaller aleatoric spatial uncertainty than the occluded corners. For instance, the red scores in Fig.~\ref{fig:case_study0} and Fig.~\ref{fig:case_study1} represent the sum of the spatial uncertainty of the vehicles' corners that face towards the Lidar sensor, and blue scores represent the uncertainty of occluded corners. These values were calculated by summing the observation noises $\sigma^2$ of the front corners and back corners separately. For all detections in Fig.~\ref{fig:case_study0} and Fig.~\ref{fig:case_study1}, the aleatoric spatial uncertainties of occluded corners were consistently higher than the ego-vehicle facing corners.
\section{\textbf{Conclusions and Discussions}} 
\label{sec:conclusion}
We have presented a probabilistic Lidar 3D vehicle detection network that captures reliable uncertainties in vehicle recognition and 3D bounding box regression tasks. Our proposed network models the epistemic uncertainty, i.e. the model uncertainty to describe data, by performing predictions several times with dropout. Our network also models the aleatoric uncertainty, i.e. the observation noises inherent in Lidar by adding an auxiliary output layer.

Epistemic and aleatoric uncertainties behave differently from each other. Experimental results showed that the epistemic uncertainty is associated with the detection accuracy. The network showed high epistemic uncertainty with samples that were different from the training dataset, such as ghost objects, big vehicles, or vehicles with abnormal bounding box regressions. Since the epistemic uncertainty displays the limitations of the vehicle detection network, it is highly valuable to be applied to efficiently querying the unseen samples to improve the model during the offline training phase (active learning). For example, when a vehicle detector trained on highways is deployed to urban areas, it is necessary to adapt the object detector in this new environment. By employing the epistemic uncertainty, the vehicle detector can be efficiently improved by actively querying objects such as pedestrians and cyclists which do not exist on highways.

Conversely, experiments showed that the aleatoric uncertainty is influenced by the detection distance and occlusion rather than detection accuracy, as distant vehicles or the occluded parts of vehicles contain high observation noises. In this way, the aleatoric uncertainty shows the sensor limitations and can be applied to improve the tracking of a vehicle position. Furthermore, we have showed that modeling the aleatoric uncertainty improved the detection performance by $1\%-5\%$, indicating that it increased the model robustness to noisy data. Finally, computing the aleatoric uncertainty of a sample requires only one-time inference. Thus, modeling aleatoric uncertainty is useful for online deployment.

One limitation of our method is the computation cost when extracting the epistemic uncertainty, where the network needs to perform multiple feed-forward passes with dropout ($0.3$ fps on a Titan X GPU). This makes epistemic uncertainty infeasible for online autonomous driving. Finding the trade-off between the performance of epistemic uncertainty and the number of feed-forward passes is an open question for the further research. 

In the future, we intend to explore uncertainties in different Lidar based object detection architectures such as one-stage detection pipeline \cite{liu2016ssd}. Furthermore, we plan to investigate more factors that may influence aleatoric uncertainties, such as Lidar reflection rates and different bounding box encodings. Finally, we plan to apply our uncertainty estimation to active learning and object tracking to improve bounding box predictions, or to incorporate these uncertainty estimations as additional knowledge to the training phase (e.g. \cite{Hu_2018_CVPR}).

\section*{Acknowledgment}
We thank Zhongyu Lou and Florian Faion for their suggestions and inspiring discussions.

\bibliographystyle{IEEEtran}
\bibliography{bibliography}

\end{document}